\documentclass[english]{cfasta}
\usepackage{booktabs}
\usepackage{multirow}
\usepackage{threeparttable}
\usepackage{makecell}
\usepackage[comma,numbers,square,sort&compress]{natbib}
\usepackage{epstopdf}
\usepackage{algorithm,multirow,xcolor}
\usepackage{algorithmicx}
\usepackage{algpseudocode}
\usepackage{subfigure}
\usepackage{amsmath}
\usepackage{amsthm, amssymb}
\usepackage{gbt7714new}

\begin{document}

\title{
	 Multi-Robot Motion Planning: A Learning-Based Artificial Potential Field Solution}

\author{
	Dengyu Zhang,
    Guobin Zhu,
    Qingrui Zhang$^*$
    }


\affiliation{
	School of Aeronautics and Astronautics, Sun Yat-sen University, Shenzhen 518107, P.~R.~China 
        \email{zhangdy56@mail2.sysu.edu.cn,
        zhugb@mail2.sysu.edu.cn, 
        zhangqr9@mail.sysu.edu.cn} }

\maketitle

\begin{abstract}

    Motion planning is a crucial aspect of robot autonomy as it involves identifying a feasible motion path to a destination while taking into consideration various constraints, such as input, safety, and performance constraints, without violating either system or environment boundaries. This becomes particularly challenging when multiple robots run without communication, which compromises their real-time efficiency, safety, and performance. In this paper, we present a learning-based potential field algorithm that incorporates deep reinforcement learning into an artificial potential field (APF). Specifically, we introduce an observation embedding mechanism that pre-processes dynamic information about the environment and develop a soft wall-following rule to improve trajectory smoothness. Our method, while belonging to reactive planning, implicitly encodes environmental properties. Additionally, our approach can scale up to any number of robots and has demonstrated superior performance compared to APF and RL through numerical simulations. Finally, experiments are conducted to highlight the effectiveness of our proposed method.

\end{abstract}

\footnotetext{This work is supported in part by the National Nature Science Foundation of China under Grant 62103451 and Shenzhen Science and Technology Program JCYJ20220530145209021 ($^{*}$Corresponding author).}

\section{Introduction}

Multi-robot systems are more efficient and flexible than single robots, and have attracted extensive researches in many applications, such as package delivery and search-and-rescue operation \cite{wurm2008IROS, bai2018RAS}.
Motion Planning is a fundamental module of robot autonomy, which computes dynamically-feasible trajectories for robots to reach their perspective goal regions \cite{mohanan2018RAS}. 
Yet, motion planning is a challenging task in cluttered environments for multiple robots. 
It's demanding to find a safe and smooth motion towards a target in a congested environment. 
Also, in the same space, robots need to consider motions of others, while communications among them are scarcely available\cite{kretzschmar2016IJRR}. 
The decision difficulty, planning cost, and collision risk grow exponentially with the amount of robots and obstacles in cluttered environments.

Most motion planning methods can be summarized into two categories, namely proactive and reactive algorithms. 
Proactive planning makes active decisions based on both current and forthcoming states of the environment \cite{Sandeep2019IJRR, Alonso2018TRO}. 
Most proactive planning algorithms are based on online optimization, \emph{e.g.}, dynamic window approach (DWA), model predictive control (MPC) \cite{zhu2019IRAL, soria2021Nature}, \emph{etc}. 
Lightweight platforms, such as UAVs and UGVs, can hardly afford cumbersome computation of those optimization-based algorithms. 

Reactive planning determines motion directly based on the instantaneous measurements from on-board sensors \cite{arslan2019IJRR}. In reactive planning,
real-time motion is generated as a function of a certain vector field or virtual forces, \emph{e.g.}, Dipole-like field \cite{panagou2014ICRA}, guiding vector field \cite{Yao2021TRO}, and artificial potential field (APF) \cite{ZhangZheng2022}, \emph{etc}. 
Although solutions are usually not optimal, reactive planning algorithms are more computationally efficient than optimization-based proactive algorithms. 
Hence, reactive planning is suitable for lightweight platforms with limited resources, but robot motion by  reactive planning might be oscillatory.

In this paper, a learning-based potential field method is proposed for distributed multi-robot motion planning. 
Reinforcement learning (RL) is introduced to enhance the conventional artificial potential field. 
An observation embedding is used to gather environment information into a fixed-length vector. 
With the embedded real-time information, RL module is trained to adjust parameters of APF dynamically. 
A soft wall-following rule is developed to reduce oscillations of APF, thus the reference input of inner-loop controller is determined. 
The efficiency of the proposed algorithm is evaluated extensively using numerical simulations, and also validated through experiments with quadrotors. 
In summary, the main contribution of this paper are listed below:
\begin{enumerate}
	\item A distributed reactive planning framework is developed for multiple robots in cluttered environments. The proposed design integrates RL into conventional APF, and is able to make active responses to dynamic change of surrounding environments. 
	\item An observation embedding is integrated, which is able to extract the hidden features of the environments.
	\item A soft wall-following rule is presented, which could improve the smoothness of the planned trajectory.
\end{enumerate}


\section{Related Works}
\label{sec:Related}

For multi-robot motion planning problems, several proactive methods have been proposed, such as decenteralized MPC \cite{Tallamraju2018SSRR} and sequential MPC \cite{Aoki2022ITSC}. As a model-based method, MPC can use dynamic characteristics of the model to ensure a result with smoothness, stability and feasibility. However, to solve multi-robot problems by MPC, communication and computation is necessary for intent sharing and decision optimizing. When the burden of communication and computation grows exponentially with the amount of robots and obstacles, lightweight platform may be unable to support real-time planning.

Reactive planning methods are mainly derived from Artificial Potential Field (APF). APF defines two types of virtual field, namely attractive field for target reaching and repulsive field for obstacle avoidance \cite{Khatib1986APF}. APF decomposes complex planning problems into simple potential field superposition problems. Hence, APF is capable of safe navigation in sparse stationary environments at a low cost. However, local minima issue is a fundamental defect of APF, so no assurance for global convergence to target location is guaranteed \cite{mohanan2018RAS}. The local minima issue can be solved in environments with sufficiently separated obstacles \cite{arslan2019IJRR}, while the issue is still open for general cluttered environments. Vector fields (VF) is another reactive method. Based on VF, extensions on avoiding local minima was \cite{Yao2021TRO, panagou2014ICRA}, but assume the  environment is fully known.

Learning-based methods have shown significant capacity in motion planning, especially RL. RL learns policies through maximizing a certain return function using data samples \cite{kretzschmar2016IJRR}. Deep value learning can be human-like when navigating in crowded dynamic environments \cite{chen2017IROS}. Proximal policy optimization (PPO), a RL algorithm, can be used to train a sensor-level decentralized collision-avoidance policy that has an end-to-end feature \cite{Tingxiang2020IJRR}. Learning based methods may have poor generalization when details are not well designed, hence, an integrated design of learning-based and field-based methods is proposed for distributed motion planning in dynamic dense environments \cite{Semnani2020IRAL}. In our previous work \cite{ZhangZheng2022}, we integrated APF into Dueling Double Deep Q Network (D3QN) to boost the policy learning for the multi-robot cooperative pursuit problem. In this paper, a new integrated design of APF and PPO is proposed to solve motion planning problems.

\section{Preliminaries}
\label{sec:preli}

\subsection{Problem Formulation}\label{sec:problem_formulation}
The motion planning problem for $N (N\geq1)$ robots is designed as follow:
\begin{itemize}
	\item The objective for each robot is to navigate to their preassigned goals efficiently while avoiding collision.
	\item All robots are in the same environment with obstacles, and the environment is partially observable for each robot. Hence, there might be conflicts for robots to directly move towards their respective goals.
	\item All robots move at a desired speed $v$. They can only change their heading angles for both reaching goals and avoiding collision.
\end{itemize}
Such task is challenging, as robot only has one input to change its motion for two tasks in cluttered multi-robot environments.

This paper focuses on upper-level motion planning, a well-designed inner-loop controller is assumed to be implemented on the robot. Define the total set of robots as $\mathcal{V}=\left\{1,\;2\,\ldots,N\right\}$, a first-order point-mass dynamics for a robot $\forall$ $i\in\mathcal{V}$ is considered as below. 
\begin{equation}
	\dot{\boldsymbol{p}}_i = \boldsymbol{v}_i 
	\label{eq:SysDyn}
\end{equation}
where $\boldsymbol{p}_i\in\mathbb{R}^m$ is the robot position vector with $m\geq2$, and $\boldsymbol{v}_i\in\mathbb{R}^m$ is the velocity vector.

\begin{figure}[tbp]
	\centering        
	\includegraphics[width=0.6\linewidth]{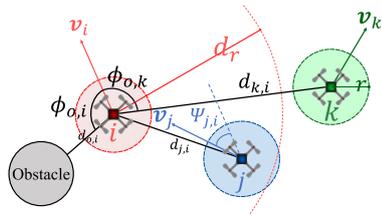}
	\caption{Problem background (All robots are homogeneous with the same safe radius dented by $r$. A robot $i$ is able to detect obstacles and neighboring robots in a range of $d_r$. Necessary exteroceptive sensors are mounted on the robot $i$ to obtain surrounding information, such as the obstacle distance $d_{o, i}$, the azimuth angle of an obstacle, the distance of a neighboring robot $d_{j, i}$,  azimuth angle of a neighboring robot $\phi_{j,i}$, and the relative heading angle $\psi_{j,i}$ between robot $i$ and its neighbor $j$.).}
	\label{fig:problem_formulation}
\end{figure}

As shown in Fig. \ref{fig:problem_formulation}, all robots of interest in the environment are homogeneous, which has a safe radius of $r$.  Let $d_{o, i}$  be the distance from the robot $i$ to the surface of an obstacle, and $d_{j, i}= \|\boldsymbol{p}_j-\boldsymbol{p}_i\|$ be the distance between the robots $j$ and $i$ as in Fig. \ref{fig:problem_formulation}. Collisions will occur if $d_{o, i}<r$ or $d_{j,i}<2r$. We use $\phi_{o,i}$ to denote the azimuth angle of an obstacle in the local frame of the robot $i$ as shown in Fig. \ref{fig:problem_formulation}. Similarly, $\phi_{j,i}$ is the azimuth angle of a robot $j$ in the local frame of the robot $i$. The relative heading angle of a robot $j$ to robot $i$ is specified by $\psi_{j,i}$. The detection or sensing range is represented by $d_r$ for each robot as illustrated in Fig. \ref{fig:problem_formulation}. An robot $i$ is able to obtain all necessary environment states in its detection range for planning, including obstacle information \{$d_{o, i}$, $\phi_{o,i}$\} and neighboring robot information \{$d_{j,i}$, $\phi_{j,i}$, $\psi_{j,i}$\}. $d_{g,i}$ is the distance from robot $i$ to its goal position. The objective of this paper is formulated as 
\begin{equation} \label{eq:Obj}
	\left\{\begin{array}{l}
		\lim_{t\to\infty} d_{g,i}(t)<r \text{ with }  d_{g,i}(0)>0   \\
		d_{j,i}(t)>2r \quad \forall j\in\mathcal{V}, j \neq i \\
		d_{o,i}(t) > r
	\end{array}
	\right.
	\forall\; i\in \mathcal{V}
\end{equation}

\subsection{Reinforcement Learning}\label{sec:RL}
In RL, state transitions of environments are formulated as a Markov Decision Process (MDP) that is denoted by a tuple
$(\mathcal{S},\mathcal{A},\mathcal{P},{R},\gamma)$, where
\begin{itemize}
	\item $\mathcal S$ is the state space with $s_t \in \mathcal S$ denoting the environment state at the timestep $t$.
	\item $\mathcal A$ is the action space with $a_t \in \mathcal A$ being the action executed at the timestep  $t$.
	\item $\mathcal{P}$ is the state-transition probability function.
	\item ${R}$ is the reward function; 
	\item $\gamma \in \left(0,1\right)$ is a discount factor balancing the instant reward with future reward.
\end{itemize}

\begin{figure*}[htbp]
	\centering
	\includegraphics[width=0.9\linewidth]{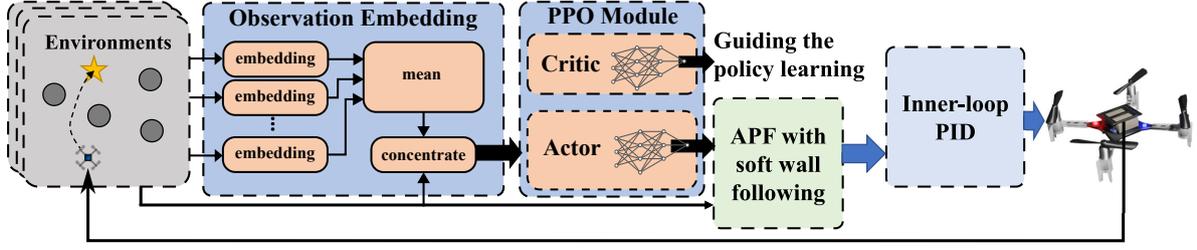}
	\caption{The overview of the learning-based potential field method. Observations are transformed to a fixed-length high-dimensional vector by an embedding. The whole network will be optimized by using a PPO agent. Hence, the PPO agent is used to encode environment information that will be used later to regulate an APF with soft wall following.}
	\label{fig:Framework}
\end{figure*}
Let the policy in RL be $\pi(a|s) : \mathcal{S}\times \mathcal{A} \to \left[0\; 1\right]$. The objective is to maximize the following accumulated return.
\begin{equation}
	\mathbb{E} [\mathcal{R}_t]= \mathbb{E}_{a\sim\pi, s\sim\mathcal{P}}[\sum_{t=k}^T\gamma^{t-k} R_t] \label{eq:ReturnRt}
\end{equation}
where $a_t\sim\pi(a_t|s_t)$, $s_{t+1}\sim \mathcal{P}(s_{t+1} | s_t,a_t)$, and $T$ is the task horizon.
The objective of RL is to learn an optimal policy $\pi^*(a|s)$ that maximizes the expectation of accumulated rewards $\mathbb{E}[\mathcal{R}_t]$ in \eqref{eq:ReturnRt}. The maximization problem can be resolved using diverse methods. In this paper, we adopt the well-known proximal policy optimization (PPO) algorithm \cite{schulman2017proximal}. Assume that the policy $\pi(a|s)$ is approximated using a deep neural network with the parameter set $\theta$, so the parameterized policy is denoted by $\pi_\theta(a|s)$. In PPO, the policy parameter is optimized by maximizing a surrogate objective as below. to limit the update of parameters into the trust region as follows \cite{schulman2017proximal}.
\begin{equation}  \label{eq:SurrObj}
	\begin{array}{ll}
		L^{CLIP} =& \mathbb{E}\left[\min\left( \frac{\pi_\theta(a_t|s_t)}{\pi_{\theta_{old}}(a_t|s_t)}A_t, \right.\right.\\
		& \left.\left.{\rm clip}(\frac{\pi_\theta(a_t|s_t)}{\pi_{\theta_{old}}(a_t|s_t)}, 1-\epsilon, 1+\epsilon)A_t\right)\right]
	\end{array}
\end{equation}
where $A_t$ is the advantage function and $\epsilon$ is the clip parameter. More details can be found in \cite{schulman2017proximal}.

\subsection{Artificial Potential Field}\label{sec:APF}
Artificial potential field methods guide robots to their predefined goals by the combination of multiple forces. The attractive force exerted on robot $i$ is defined as follows.
\begin{equation}
	\boldsymbol{F}_{a,i} = \frac{\boldsymbol{p}_{g,i} - \boldsymbol{p}_i}{d_{g,i}} \label{eq:AttrForce}
\end{equation}
Accordingly, in influence range of obstacles ($d_{o,i} \leq \rho$), the repulsive force from an obstacle is
\begin{equation}
	\label{eq:repulsive_force}
	\boldsymbol{F}_{r,i} = 
		\eta\left( \frac{1}{d_{o,i}} - \frac{1}{\rho} \right) \frac{\boldsymbol{p}_i - \boldsymbol{p}_{o,i}}{d_{o,i}^3}
\end{equation} \label{eq:RepForce}
where $\eta$ is the scale factor. $\boldsymbol{p}_{o,i}$ is the position of the nearest obstacle to robot $i$. $\rho$ is the influence range of obstacles.

The inter-robot force that keeps robots away from each other is  
\begin{equation} \label{eq:individual_force}
	\boldsymbol{F}_{in,i} = \sum_{j\neq i, j \in \mathcal{N}_i} \left( 0.5 - \frac{\lambda}{d_{j,i}} \right) \frac{\boldsymbol{p}_j-\boldsymbol{p}_i}{d_{j,i}}
\end{equation}
where $\mathcal{N}_i$ is the set of robots that can be detected by robot $i$, namely $\mathcal{N}_i=\left\{j\vert\; \forall\;j\in\mathcal{V}\text{, }j\neq i\text{, \& }d_{j,i}<d_r\right\}$, $\lambda$ is a parameter adjusting the compactness of the multi-robot system. The smaller $\lambda$ is, the more 
compact the system is \cite{ZhangZheng2022}. 
The resultant force, therefore, is 
\begin{equation}
	\boldsymbol{F}_i = \boldsymbol{F}_{a,i} + \boldsymbol{F}_{r,i} + \boldsymbol{F}_{in,i}
\end{equation}
The reference heading angle for robot $i$ is determined by $\boldsymbol{F}_i$.

\section{Algorithm}\label{sec:alg}
This paper presents a learning-based potential field algorithm for multi-robot motion planning. The algorithm integrates observation embedding, PPO, and APF into a single framework, as shown in Figure \ref{sec:alg}. The observation embedding transforms observations into fixed-length vectors, which are then passed through a Multiple-Layer Perceptron (MLP) network to generate options for a modified APF with a soft wall-following mechanism.

\subsection{Observation Embedding}

The observations of each robot is firstly split to two parts, namely $\boldsymbol{o}_{i}=\left\{\boldsymbol{o}_{loc,i},\boldsymbol{o}_{nei,i}\right\}$. The first part $\boldsymbol{o}_{loc,i}$ includes the relative distance and azimuth angles of the nearest obstacle and the goal information, so $\boldsymbol{o}_{loc,i}=\left[d_{o,i}, \phi_{o,i},d_{g,i},\phi_{g,i}\right]^T$. The second part $\boldsymbol{o}_{nei,i}$ consists of the information of observed neighbors $\boldsymbol{o}_{nei,i}=\left\{\boldsymbol{w}_j|\forall\; j \in \mathcal{N}_i\right\}$, where $\boldsymbol{w}_j$ is
\begin{equation}\label{eq:NeighInf}
	\boldsymbol{w}_j = \left[d_{j,i},\phi_{j,i}, \psi_{j, i}\right]^\top
\end{equation}

The mean embedding then employs a one-layer fully-connected networks to encode $\boldsymbol{w}_j$ into a fixed-length high-dimensional vectors $\boldsymbol{e}_j$, respectively. The networks use ReLU activation function.
\begin{equation}
		\boldsymbol{e}_j = \phi_e(\boldsymbol{o}_{loc,i}, \boldsymbol{w}_j)
\end{equation}
The information of each observed neighbors is averaged.
\begin{equation}
	\boldsymbol{c}_i = \frac{1}{|\mathcal{N}_i|}\sum_{j \in \mathcal{N}_i}{\boldsymbol{e}_j}
\end{equation}
Finally,  $\boldsymbol{o}_{loc,i}$ is concatenated with $\boldsymbol{c}_i$ to represent the observations of robot $i$ with a fixed-length vector $\hat{\boldsymbol{o}}_i$. 

\subsection{Optimization by PPO}

The action space has two dimensions corresponding to $\eta$ and $\lambda$, respectively. The range of candidate action pairs is $[0,0.1]\times[0,5]$, which is empirical selected according to the specific scenarios used in Section \ref{sec:exp}. All robots share the same policy network, so the same reward function is used for all robots.  The total reward for each robot is 
\begin{equation}
	R = R_{m} + R_{s} + R_{c} + R_{o} + R_{p}
\end{equation}
The reward $R_m$ encourages robots to reach their goals via the shortest trajectory. For a robot $i$ when it has reached its goal ($d_{g,i}<r$), one has
\begin{equation}
	R_m = 300 - 100\frac{d_{a}}{d_{s}}  
\end{equation}
where $d_{s}$ is the Euclidean distance between the start point and the goal position, and $d_{a}$ denotes the trajectory length achieved by the robot $i$. 
$R_{s}$ encourages robots to move smoothly by giving a punishment of $-5$ when the difference of headings at two adjacent timesteps exceeds $45^\circ$. 
$R_{o}$ gives a punishment of $-100$ when a robot collide with neighbors. 
$R_c$ is used to prevent robots from collisions with obstacles. For a robot $i$, one has 
\begin{equation}
	R_o = \left\{
	\begin{aligned}
		-100,\quad &{\rm if }\  d_{o,i} < r\\
		-20,\quad &{\rm if }\  r \leq d_{o,i} < 2r\\
		0,\quad &{\rm otherwise}
	\end{aligned}
	\right.
\end{equation}
The reward $R_p$ provides a dense reward to accelerate the learning process. For a robot $i$ not far from its goal ($d_{g,i} < d_m$), one has 
\begin{equation}
	R_p = 1 - \frac{d_{g,i}}{d_m}
\end{equation}
where $d_{g,i}$ is the relative distance of the goal, and $d_m>0$ is a hyperparameter. 

\begin{algorithm}[htbp]
	\caption{Multi-robot Reinforced Potential Field}
	\label {Alg:1}
	\begin{algorithmic}[1]
		\State {Algorithm initialization}
		\For {episode = $1,2,\cdots,$}
		\State {Reset the environment}
		\For {$t=1,2,\cdots,$}
		\For{$i = 1,2,\cdots, N$ }
		\If {$d_{g,i}>r$ }
		\State {Obtain $(\eta,\lambda)$ from $\pi_\theta(a|s)$}
		\State {Calculate $\boldsymbol{F}_{a,i},\boldsymbol{F}_{r,i},\boldsymbol{F}_{in,i}$ by APF}
		\If {$\boldsymbol{F}_{ar,i}^T\boldsymbol{F}_{a,i} < 0$}
		\State {Choose $n_{1,2}$ as $\boldsymbol{F}_i$}
		\Else \If {$\boldsymbol{F}_{r,i}^T \boldsymbol{F}_{a,i} < 0$}
		\State {Calculate $\boldsymbol{F}_{soft,i}$ as $\boldsymbol{F}_i$}
		\EndIf
		\EndIf
		\State {Move along the direction of $\boldsymbol{F}_i$}
		\State {Get the reward $R$ and observation $\boldsymbol{o}_i$}
		\EndIf
		\EndFor
		\If {$t \ \mathrm{mod} \ Z = 0$}
		\State {Update the PPO agent for $K$ epochs}
		\EndIf
		\EndFor
		\EndFor
	\end{algorithmic}
\end{algorithm}

To extend the classical PPO algorithm to multi-robot settings, the parameter sharing technique is employed. That is, all robots share the same policy network. The shared policy network updates parameters using transitions collected by all robots. The overall procedure of the proposed algorithm is shown in Algorithm \ref{Alg:1}.  

\subsection{Soft wall following}
\begin{figure}[htbp]
	\includegraphics[width=\linewidth]{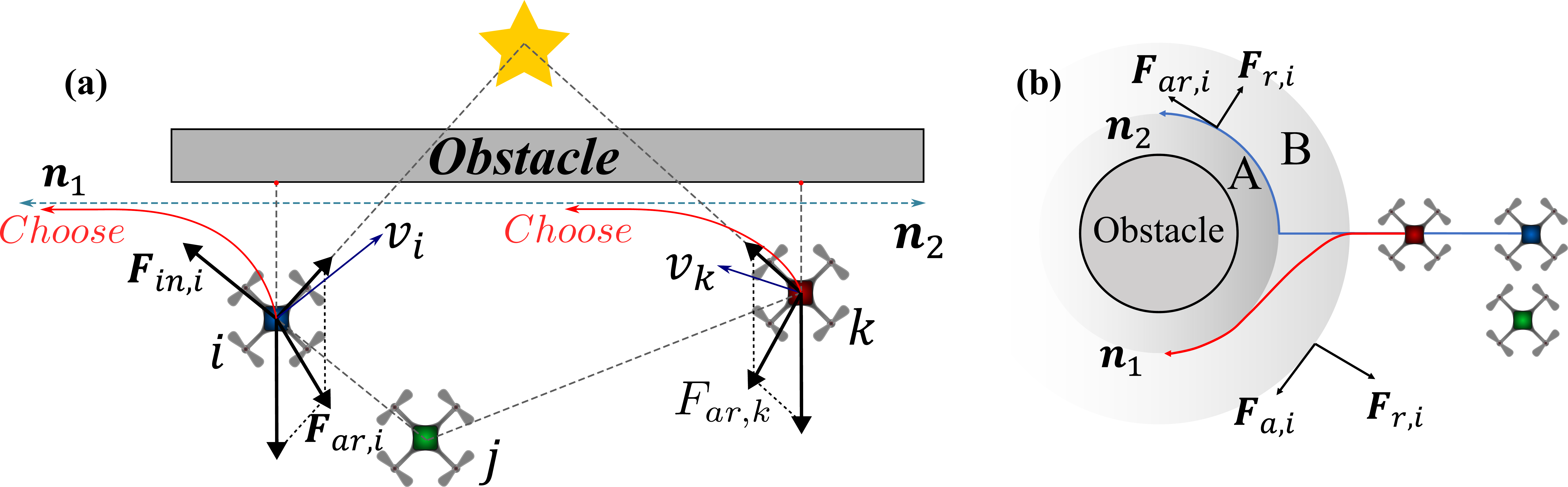}
	\caption{Wall following rule. (a): agent $i$ (blue), agent $j$ (green), agent $k$ (red) are moving around an obstacle. The soft extension of wall following is shown in (b). In area A, the angle between the resultant force $F_{ar,i}$ and the attractive force $F_{a,i}$ exceeds 90°, where wall following rule is active. In Area B, the angle between the attractive $F_{a,i}$ and the repulsive $F_{r,i}$ exceeds 90°, where robots move according to soft wall following rule.}
	\label{fig:wall-following}
\end{figure}

In order to tackle the local minimum problem of ApF, a wall following rule is  introduced, which allows robots to move along the edge of obstacles. As shown in Fig \ref{fig:wall-following}, $\boldsymbol{n}_1$ or $\boldsymbol{n}_2$ is the better direction for robots rather than $\boldsymbol{F}_{ar, i} =\boldsymbol{F}_{a,i}+\boldsymbol{F}_{r,i}$.
The wall following rule is activated when the angle between $F_{ar,i}$ and  $F_{a,i}$ exceeds $90^{\circ}$. 
The selection of $\boldsymbol{n}_1$ and $\boldsymbol{n}_2$ depends on the current heading direction and inter-robot force $\boldsymbol{F}_{in,i}$.
If $\boldsymbol{F}_{in,i}$ exceeds the specified a threshold $\bar{F}_{in,i}$, we select from $\boldsymbol{n}_1$ and $\boldsymbol{n}_2$ the one that has a smaller angle with $\boldsymbol{F}_{in,i}$ as shown in Fig. \ref{fig:wall-following} (a). Otherwise, choose the one with a smaller angle with the current motion direction as the robot $k$ in Fig. \ref{fig:wall-following}.

When the angle between $\boldsymbol{F}_{ar,i}$ and $\boldsymbol{n}_{1}$ (or $\boldsymbol{n}_{2}$) is very large, there might be a sharp turn as the blue trajectory shown in Fig. \ref{fig:wall-following}(b). To solve the problem, we introduce a soft rule to the wall following method. The nearby external area of an obstacle is divided into two sub-areas as shown in Fig. \ref{fig:wall-following}(b). In the sub-area A, a robot $i$ chooses $\boldsymbol{n}_1$ or $\boldsymbol{n}_2$ according to the aforementioned wall following rule.  In the sub-area B, the robot chooses a direction $\boldsymbol{F}_{soft}$ between $\boldsymbol{n}_{1\text{ or }2}$ and $\boldsymbol{F}_{ar,i}$. Hence, $\boldsymbol{F}_{soft,i}$ is defined as
\begin{equation}
	\boldsymbol{F}_{soft,i} = \frac{\boldsymbol{F}_{ar,i} + 2\|\boldsymbol{F}_{r,i}\|  \boldsymbol{n}_{1\text{ or }2}}{\|\boldsymbol{F}_{ar,i} + 2\|\boldsymbol{F}_{r,i}\|  \boldsymbol{n}_{1\text{ or }2}\|}
\end{equation}

At the exact moment that a robot $i$ enters the sub-area B, $\boldsymbol{F}_{soft,i}$ is the same as $\boldsymbol{F}_{ar,i}$. When agents get closer to obstacle, $\boldsymbol{F}_{soft,i}$ becomes closer to $\boldsymbol{n}_{1\text{ or }2}$.

The sub-area A and B around an obstacle in Fig \ref{fig:wall-following} are defined as follow: If $\boldsymbol{F}_{ar,i} ^T \boldsymbol{F}_{a,i} < 0$, it belongs to sub-area A; if $\boldsymbol{F}_{r,i}^T \boldsymbol{F}_{a,i} < 0$, it is in sub-area B.

\begin{figure}[htbp]
	\centering
	\includegraphics[width=0.9\linewidth]{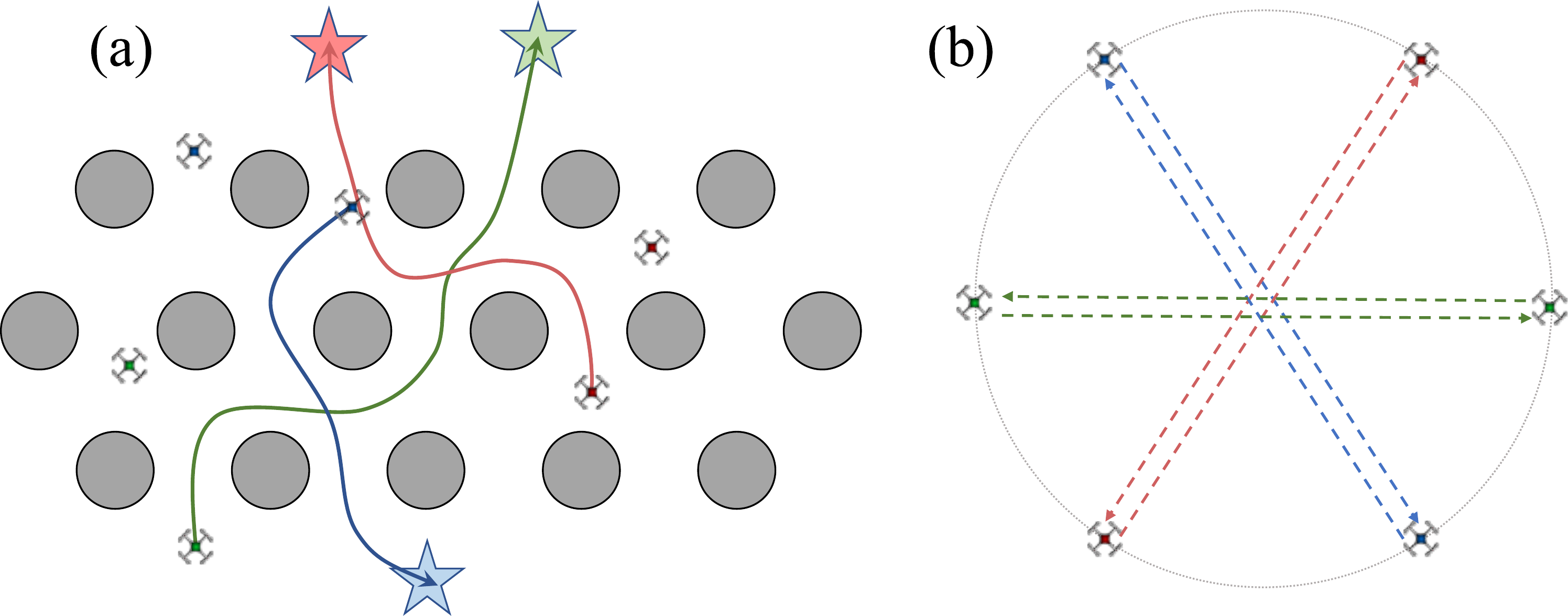}
	\caption{Arena for algorithm training. (a) Six robots go to their destinations, while avoiding dense obstacles (gray). The start point and goal positions of robots are generated randomly. (b) Six robots swap their locations with one another with random initial positions.}
	\label{fig:train_scene}
\end{figure}

\begin{table}[htbp]
	\centering
	\caption{Training setup}
	\label{tab:train_params}
	\begin{tabular}{c c}
		\toprule
		Parameters & Values \\
		\midrule
		
		Actor network &  MLP with 2 hidden layers \\
		&(256 neurons per hidden layer) \\
		Critic network &  MLP with 2 hidden layers \\
		&(256 neurons per hidden layer) \\
  Learning rate $\alpha_0$ & 0.0003 \\
		$\beta$ & 0.999 \\
  $\gamma$ & 0.999 \\
  $c_2$ & 0.001 \\
  Time steps per episode & 1000 \\
  Training episodes & 1000 \\
		Batch size $Z$ & 100 \\
  $\epsilon$ & 0.2 \\
		$\tau$ & 0.9 \\
		$c_1$ & 0.5 \\
        $K$ & 1 \\
        \bottomrule
	\end{tabular}
\end{table}

\section{Simulation and experiments}
\label{sec:exp}
In this section, numerical simulation and experimental results are presented. 
The proposed algorithm is trained in two different arenas with six robots as shown in Fig. \ref{fig:train_scene}. 
In the first arena, the environment is occupied with many obstacles as shown in Fig. \ref{fig:train_scene} (a). 
The obstacles have the same radius of $0.5$ $\mathrm{m}$. 
Both the initial and goal positions of each robot are randomly given. 
The second training arena is shown in Figure \ref{fig:train_scene} (b). 
Initially, all robot are distributed evenly on a circle with a radius of $2$ $\mathrm{m}$. 
Robots are required to swap their locations. 
In the training, the desired speed of agents is set to $0.5$ $\mathrm{m/s}$. 
The safety radius $r$ of robots is $0.1$ $\mathrm{m}$. 
The perception range $d_p$ of a robot is chosen to be $6$ $\mathrm{m}$. 
The time step at training is set to be $0.1$ $\mathrm{s}$. 
The hyperparameter $d_m$ is chosen to be $10$ $\mathrm{m}$. 
The influence range $\rho$ of an obstacle is chosen $10$ $\mathrm{m}$. 
The wall following threshold $\bar{F}_{in,i}$ is picked to be $1$, $\forall$ $i\in\mathcal{V}$. 
The learning rate decays as follows. 
\begin{equation}
	\alpha = \alpha_0 \times \beta ^ {j}
\end{equation}
where $\alpha_0$ is the initial learning rate,  $\beta$ is a hyperparameter, and $j$ is the number of the current episode. The values of $\alpha_0$, $\beta$ are given in Table \ref{tab:train_params}. 

The proposed RPF is compared with two different methods, namely vanilla PPO and vanilla APF. For vanilla PPO, we found that it's very difficult to train a policy in the cluttered environment shown in Fig \ref{fig:train_scene} (a). For comparison, we train the vanilla PPO policy in an environment with smaller obstacles of radii $0.1 \mathrm{m}$. The locations of the obstacles are the same with those in Fig \ref{fig:train_scene} (a). Other learning configurations are the same with those as RPF as given in Table \ref{tab:train_params}. The decision of PPO robot is generated as
\begin{equation}
	\boldsymbol{F}_i = \frac{\boldsymbol{v}_i + a_t \boldsymbol{v}_{i\perp}}{\|\boldsymbol{v}_i + a_t \boldsymbol{v}_{i\perp}\|}
\end{equation}
where $\boldsymbol{v}_{i\perp}$ is a vector perpendicular to $\boldsymbol{v}_i$, and $a_t$ is the output by the PPO policy with $a_t\in[-2.5, 2.5]$.

We choose the parameters of a vanilla APF to be $\eta=0.05$ and $\lambda=2$. For fair comparison, we choose two metrics to evaluate the performance of all methods in the simulation. 

The \textbf{traveling distance} metric $l_i$ evaluates the average total traveling length by a robot, which is defined as below.
\begin{equation}
	l_i = \sum_{i \in \mathcal{V}}\sum_{t}^T ||\boldsymbol{v}_i(t)||\Delta t
	\label{eq:distance}
\end{equation}
where $T$ represents the total timesteps for robots to reach their goal. The large the traveling distance metric is, the poor the algorithm performance is. 

The \textbf{motion smoothness} metric $\xi_i$ evaluates how oscillatory the motion of a robot, which is defined as 
\begin{equation}
	\xi = \frac{\sum_{i\in\mathcal{V}} \sum_t^T ||\Delta \boldsymbol{v_i}|| / ||\boldsymbol{v_i}||}{T}
	\label{eq:change_angle}
\end{equation}
Similarly, the large the motion smoothness metric is, the poor the performance of an algorithm is. 

\subsection{Simulation Evaluation} \label{subsec:sim}
\begin{figure}[htbp]
	\centering
	\includegraphics[width=\linewidth]{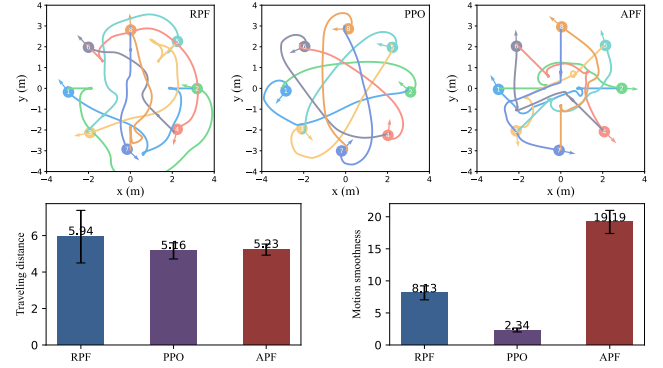}
	\caption{Simulation evaluation in an open environment. All $8$ robots are initially distributed on a circle with a radius of $3$ $\mathrm{m}$ and swap their locations.
	}
	\label{fig:compare_circle}
\end{figure}

In the simulation evaluate, the trained proposed algorithm is compared with the other two algorithms to show its efficiency and generalization. In the two scenarios, open environments with no obstacles are considered. All simulations and comparisons are given in Fig. \ref{fig:compare_circle}. Both the traveling distance and motion smoothness metrics in Figs. \ref{fig:compare_circle} are averaged among all robots to evaluate the whole performance of the algorithms in a multi-robot setting.

In the first evaluation, we consider $8$ robots distributed on a circle with a radius of $3$ $m$ and swap locations with robots on the other side. This scenario is very similar to our second training arena but with a different number of robots and a different size of circle. The robot trajectories are shown in Fig. \ref{fig:compare_circle}. In this scenario, PPO has the best performance in by both the traveling distance metric and the motion smoothness metric as shown in Fig. \ref{fig:compare_circle}. The APF method has the worst performance by both metrics.

\begin{figure}[htbp]
	\centering
	\includegraphics[width=0.98\linewidth]{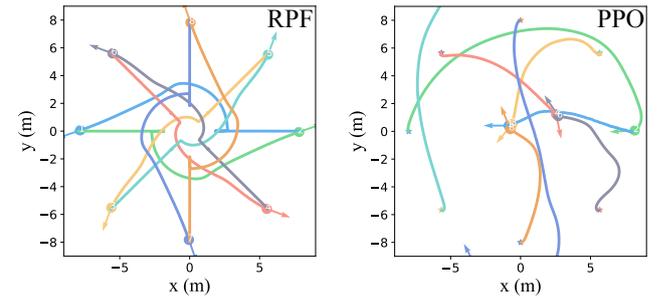}
	\caption{Comparison of RPF and PPO with $8$ robots initially distributed on a circle with a radius of $8$ $\mathrm{m}$.
	}
	\label{fig:ppo_fail}
\end{figure}
Although the vanilla PPO has the best performance in the first scenario, it cannot generalize well to a scenario with more difference as shown in Figs. \ref{fig:ppo_fail}. In the second evaluation in Figs. \ref{fig:ppo_fail}, we further compare our method RPF with the vanilla PPO in a more different scenaro. In this case, $8$ robots are initally distributed on a circle with a radius of $8$ $\mathrm{m}$ that is large than the detection range ($6$ $\mathrm{m}$) of a robot. The same position swap task is performed. In such a scenario, the vanilla PPO failed to complete the task safely.

\subsection{Experiment Evaluation}
\begin{figure}[htbp]
	\centering
	\subfigure{
		\includegraphics[width=0.9\linewidth]{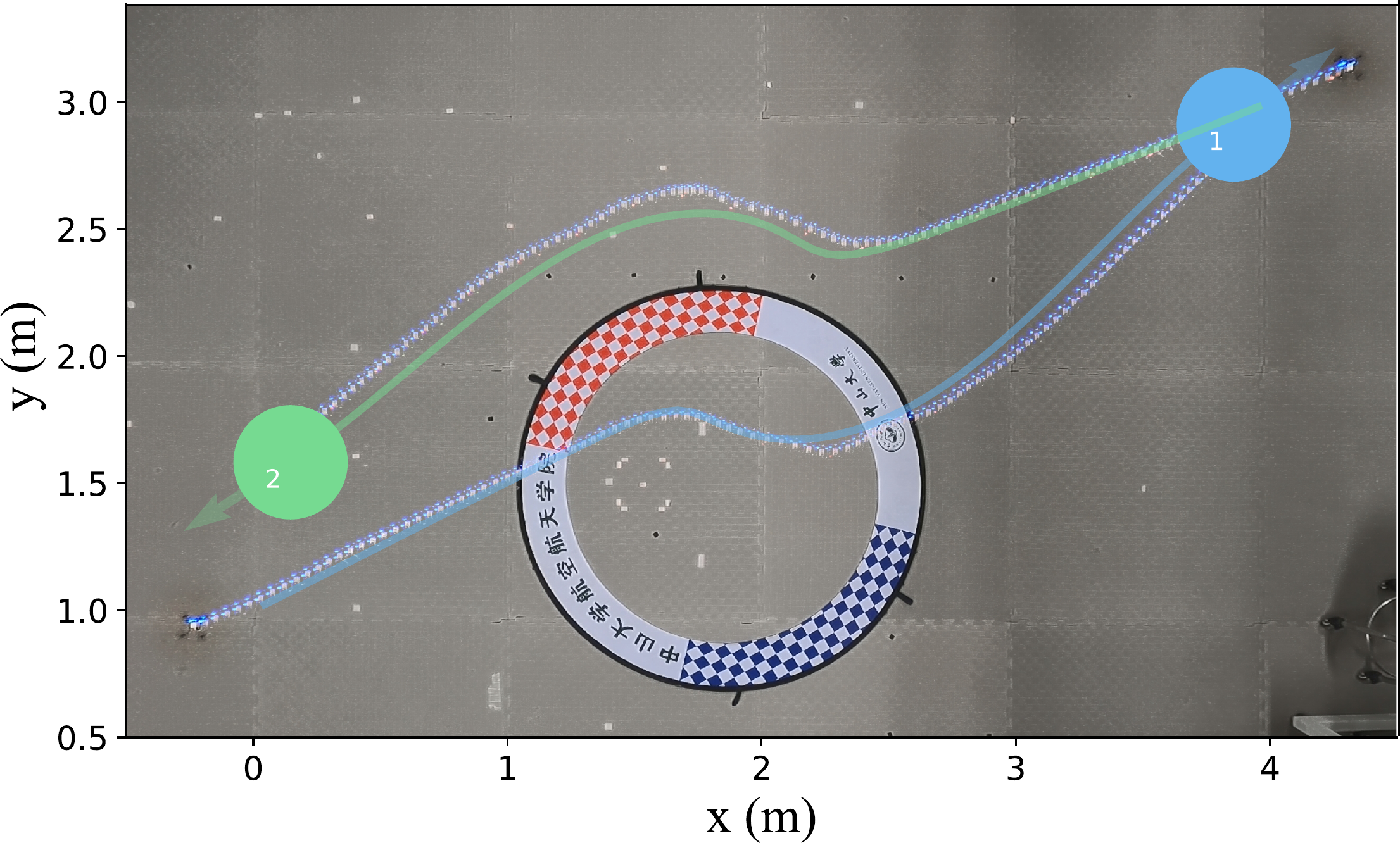}
	}
	\caption{An experiment that two drones swap their positions avoiding collision with others. The border of experiment region is treated as an rectangle obstacle. Simulation corresponding to the experiment is provided.}
	\label{fig:real_exp}
\end{figure}
An experiment evaluation was conducted to demonstrate the effectiveness of the proposed algorithm in a real-life platform. In the experiment, the RPF algorithm was applied to Crazyflie quadrotors without any modification. The OptiTrack optical motion capture system was used to measure the states of the quadrotors, and the implemented inner-loop controller enabled the Crazyflie drones to track position commands from the high-level RPF planner. During the experiment, two drones successfully swapped their positions while avoiding collisions with other drones. The results demonstrated that the quadrotors were able to change their decisions smoothly and resolve conflicts effectively, which is consistent with simulation results. A snapshot of the experiment can be seen in Fig \ref{fig:real_exp}.



\section{Conclusion} \label{sec:Conclusion}
This paper introduced a new motion planning algorithm that integrates deep reinforcement learning with an artificial potential field. The proposed algorithm can actively respond to dynamic changes in the environment and implicitly encodes environmental properties with an observation embedding method. Additionally, a soft wall-following rule was presented to improve path smoothness. Numerical simulations demonstrated that our algorithm outperforms existing benchmarks such as vanilla PPO and APF. Real experiments were conducted to show the effectiveness of the proposed method in real-world systems. In future work, we plan to integrate sensor observations into our approach to implement it in diverse scenarios.

\section{Acknowledgement}\label{sec:Acknow}
This work is supported in part by the National Nature Science Foundation of China under Grant 62103451 and Shenzhen Science and Technology Program JCYJ20220530145209021.

\bibliographystyle{gbt7714-numerical}
\bibliography{references.bib}

\end{document}